\documentclass{esannV2}
\usepackage[dvips]{graphicx}
\usepackage[utf8]{inputenc}
\usepackage{amssymb,amsmath,array}
\usepackage{amsthm}
\usepackage[english]{babel}

\usepackage[svgnames]{xcolor} 

\usepackage{algorithm}
\usepackage{algorithmicx}
\usepackage{algpseudocode}

\newlength{\bibitemsep}
\newlength{\bibparskip}
\let\oldthebibliography\thebibliography
\renewcommand\thebibliography[1]{%
  \oldthebibliography{#1}%
  \setlength{\parskip}{\bibitemsep}%
  \setlength{\itemsep}{\bibparskip}%
}

\theoremstyle{definition}
\newtheorem{definition}{Definition}[section]
\usepackage{enumitem}
\setlist{noitemsep,topsep=0pt,parsep=0pt,partopsep=0pt,font=\small}

\newcommand\MK{{\mathcal{K}}}

\newcommand\MH{{\mathcal{H}}}
\newcommand\MHp{{\mathcal{H}_{+}}}
\newcommand\MHn{{\mathcal{H}_{-}}}
\newcommand\K{{K}}
\newcommand\tK{{\tilde{K}}}
\newcommand\G{{G}}
\newcommand\tG{{\tilde{G}}}
\newcommand\Kp{{K_{+}}}
\newcommand\Kn{{K_{-}}}
\newcommand\Pp{{P_{+}}}
\newcommand\Pn{{P_{-}}}

\newcommand\krein{{Kre{\u{\i}}n }}
\newcommand\dbR{{\mathrm{I\hskip-2.2pt R}}}
\newcommand\tfp{{\tilde{f_{+}}}}
\newcommand\tfn{{\tilde{f_{-}}}}
\newcommand\tf{{\tilde{f}}}
\newcommand\talpha{{\tilde{\alpha}}}

\begin{document}
\title{TrIK-SVM : an alternative decomposition for kernel methods in Kre\u{\i}n spaces}

\author{Gaëlle Loosli$^{1,2}$
%
%
\vspace{.3cm}\\
%
1- PobRun \\
Brioude, France \\
%
2- UCA - LIMOS UMR 6158 CNRS  \\
Clermont-Ferrand, France\\
}

\maketitle

\begin{abstract}
The proposed work aims at proposing a alternative kernel decomposition in the context of kernel machines with indefinite kernels. The original paper of KSVM (SVM in \krein spaces) uses the eigen-decomposition, our proposition avoids this decompostion. We explain how it can help in designing an algorithm that won't require to compute the full kernel matrix. Finally we illustrate the good behavior of the proposed method compared to KSVM.
\end{abstract}
\vspace{-0.7em}
\section{Introduction} \vspace{-0.7em}
Learning with indefinite kernels is a question that has been treated many times since the introduction of the hyperbolic tangent kernel (tanh) in the libsvm toolbox \cite{hsuan-tien_study_2003}. This particular kernel has never really shown its practical interest, but many other cases of indefinite kernels have been published, such as graph kernels or similarity based kernels \cite{chen_fusing_2009}. In those cases, the need for an indefinite kernel is driven by the application field and the cost of avoiding the indefiniteness can be huge in terms of accuracy \cite{loosli_study_2016}.
On the theoretical side, \cite{ong_learning_2004} have shown that learning with an indefinite kernel implies learning in reproducing kernel Kre\u{\i}n spaces (RKKS), in which basically the scalar product between a vector and itself can be negative.
In \cite{loosli_learning_2016}, the KSVM algorithm is proven to produce a valid solution in an RKKS, even though the optmization in those spaces is not well defined.
We anchor our work in the same framework, and we propose an upgraded algorithm for KSVM, named TrIK-SVM.

\vspace{-0.7em}
\section{Essentials of RKKS}\vspace{-0.7em}
A Kre\u{\i}n space is a vector space in which the dot product of two identical vectors can provide a negative value. A well-known example of a Kre\u{\i}n space is the complex space, in which this property is denoted by $i^2 = -1$.
In the learning context, it has been shown that trying to optimize a quadratic program (for instance an SVM) using an indefinite kernel matrix is actually a stabilization problem in a Reproducing Kernel Kre\u{\i}n space \cite{ong_learning_2004,loosli_learning_2016}.Those spaces are indefinite inner product spaces endowed with a Hilbertian topology without any requirement of positive-definiteness.

\theoremstyle{definition}\small
\begin{definition}{\krein space \cite{_azizov_linear_1989}}
 An inner product space $(\MK, \langle . , . \rangle_{\MK})$ is a \krein space if there exists two Hilbert spaces $\mathcal{H}_{+}$, $\mathcal{H}_{-}$ spanning $\mathcal{K}$, with $f_{+} \in \mathcal{H}_{+}$
  and $f_{-} \in \mathcal{H}_{-}$, such that
\begin{equation}
\small
\left|
  \begin{array}{ll}
    \forall f \in \mathcal{K} & f = f_{+} \oplus f_{-} \\
    \forall f,g \in \mathcal{K} & \langle f,g\rangle_{\mathcal{K}} =  \langle f_{+},g_{+}\rangle_{\mathcal{H}_{+}} - \langle f_{-},g_{-}\rangle_{\mathcal{H}_{-}}\\
  \end{array}
\right.
\end{equation}
\end{definition}
\normalsize
\noindent In a nutshell, we provide here the essential facts that can be found in details in the above-cited literature:
\begin{itemize}
  \item If $(\MHp,\langle .,. \rangle)$ and  $(\MHn,-\langle .,. \rangle)$ are RKHS, $\MK = \MHp + \MHn$ is an RKKS.
  \item There is a representer thoerem in a RKKS, and the decomposition of the reproducing kernel $k = k_{+} - k_{-}$ consists of reproducing kernels in the associated RKHS.
  \item Any symmetric indefinite kernel can be decomposed in $k_{+}$ and $k_{-}$ and associated to an RKKS
  \item This decomposition is not unique
\end{itemize}
Solving a quadratic program in an RKKS ($\MK$) is a stabilization problem that cannot be directly solved. However, it can be reformulated as an equivalent minimization problem \cite{loosli_learning_2016} lying in an associated RKHS ($\MH$) defined as the sum of the positive ($\MHp$) and negative ($\MHn$) parts of the RKKS. The solution of the minimization problem lies in $\MH$ and thus needs to be projected back to $\MK$.
While the original paper shows the interest of KSVM considering the accuracy of the method on a large variety of problems,
 KSVM is not used in practice.
Indeed, it suffers from several major practical drawbacks: it requires to compute the eigen-decompostion of the matrix (computation time and stability issues) and the final solution is full (the sparsity is lost during the projection) back to the RKKS.
Recently, several papers dealing with indefinite matrices have noted those drawbacks and proposed alternative methods. Essentially, some paper are proposing minimization methods that ignore the stabilization problem \cite{xu_solving_2017} and others rely on using kernel as features, which is a way to embed the RKKS in an RKHS \cite{huang_indefinite_2017,mehrkanoon_indefinite_2018}. The first category of methods are likely to provide an admissible but not optimal solution in the RKKS. The second category usually works pretty well but might hide some information from the negative part. One can note a major improvement over KSVM in terms of computational complexity, brought in \cite{schleif_indefinite_2017} in the form of a carefully approximated version, based on kernel approximation, partial eigen-decomposition and CVM algorithm. The final solution is also approximated in order to be sparser \cite{schleif_sparsification_2018}.
\vspace{-0.7em}
\paragraph{The stabilization problem}
We recall here the stabilization problems in both RKKS and RKHS to exhibit the place of the kernel decomposition that will be discussed later.
Let $\{ x_i \in \mathcal{X}, y_i \in [-1,1], \forall i \in [1,\dots,\ell] \}$ be a binary training set, and $\tau$ an hyper-parameter for error penalization, the following problem aims at finding coefficients $\alpha_i (\forall i \in [1,\dots,\ell])$  and a bias $b$ such
that the decision function $D(x) = sign(f(x) + b)$ with $f(x)= \sum_{i=1}^{\ell} \alpha_i y_i k(x_i ,x)$.
The equivalent minimization problem is formulated by replacing $f$ by $\tf$, such that
$\langle \tf,\tf \rangle = \langle \tfp,\tfp \rangle_{\MHp} + \langle \tfn,\tfn \rangle_{\MHn}$ and $\tf = \tfp + \tfn$.
Then the equivalent minimization problem to eq.\ref{eq:stabK} becomes:

\vspace{-0.7em}
\begin{equation}
\label{eq:stabK}
\footnotesize
\begin{array}{ll}
\left\{
  \begin{array}{ll}
    \displaystyle \underset{\small f \in \MK, b \in \dbR}{\mbox{stab}} & \displaystyle \frac{1}{2} \langle f,f\rangle_{\MK} \\
    \mbox{s.t.} & \displaystyle \sum_{i=1}^{\ell} \max(0, \\
    & (1-y_i(f(x_i)+b)))\leq \tau
  \end{array}
\right.
&
\left\{
  \begin{array}{ll}
    \displaystyle \underset{\small \tf \in \MH, b \in \dbR}{\mbox{min}} & \displaystyle \frac{1}{2} \langle \tf,\tf\rangle_{\MH} \\
    \mbox{s.t.} & \displaystyle \sum_{i=1}^{\ell} \max(0, \\
    & (1-y_i(\tf(x_i)+b))) \leq \tau
  \end{array}
\right.
\end{array}
\end{equation}
The dual can be calculated as a usual SVM and it lets the modified kernel matrix $\tK = \Kp + \Kn$ appear :
\begin{equation}
\label{eq:stabDual}
\footnotesize
\left\{
  \begin{array}{cc}
    \displaystyle \underset{\small \talpha}{\mbox{max}} & \displaystyle -\frac{1}{2} \talpha^{\top}Y(\Kp + \Kn)Y\talpha + \talpha^{\top}\mathbf{1} \\
    \mbox{with} & \talpha^{\top}y = 0 \\
    \mbox{and} & 0\leq \talpha_i \leq C \quad \forall i \in [1\dots\ell]
  \end{array}
\right.
\end{equation}
$Y$ is the diagonal matrix of labels $y$. This problem provides a solution that needs to be projected in the \krein space and all the previous computations are {\em independent of the choice of the decomposition}.

\vspace{-0.7em}
\section{TrIK-SVM}\vspace{-0.7em}
Our new algorithm aims at solving eq.\ref{eq:stabK} in a more efficient way than KSVM. First let's detail the KSVM decomposition \cite{loosli_learning_2016}, which is naturally based on the eigen-decomposition of the kernel matrix. Then we give the new decomposition.
\vspace{-0.7em}
\paragraph{KSVM "flip" Decomposition}
Let $U$ and $D$ be respectively the eigenvectors and the diagonal matrix containing the eigenvalues ok the kernel matrix $K$, such that $K = U^{\top}DU$.
The modified semi-definite positive (SDP) kernel associated to $K$ is defined as $\tK = U|D|U^{\top}$ where $|U|$ contains the absolute values of $U$. In this case, $\Kp = U \max(D, 0)U^{\top}$ and $\Kn = -U \min(D, 0)U^{\top}$.
\vspace{-0.7em}
\paragraph{TrIK-SVM "shift" Decomposition}
Admitting we know the most negative eigenvalue of the kernel matrix, denoted $\lambda$:
\begin{equation}
\label{eq:decomp}
\footnotesize
K = (K - \lambda I) + \lambda I
\end{equation}
with $I$ the identity matrix. Checking that $\Kp = K - \lambda I$ and $\Kn = -\lambda I$ are both SDP matrices is straight forward. Then the modified SDP
kernel becomes $\tK = K - 2\lambda I$.

\vspace{-0.7em}
\paragraph{Transition matrices}
Transition matrices are used to project either the kernel matrix or the solution back and forth between the RKKS and the RKHS.
\begin{equation}
\footnotesize
\begin{array}{l|l}
KSVM  & TrIK-SVM \\
    \begin{array}{rl}
      \Pp = U_{+} U_{+}^{\top}  & \Pn =  U_{-} U_{-}^{\top} \\
      P = \Pp - \Pn & \\
       = USU^{\top} & \mbox{with } S = sign(D) \\
    \end{array}
  &
    \begin{array}{l}
      \Pp = (K-\lambda I)K^{-1} = I - \lambda K^{-1} \\
      \Pn = \lambda K^{-1}\\
      P = \Pp - \Pn = I - 2 \lambda K^{-1}\\
    \end{array}
\end{array}
\end{equation}

\vspace{-0.7em}
\paragraph{Algorithm}
Based on the transition matrix, the algorithm of KSVM is very simple, it consists in projecting the indefinite kernel matrix $K$ into the RKHS ($\tK$), solve a regular SVM and project the obtained solution $\talpha$ back to the RKKS ($\alpha$).
The most straight forward way to use the proposed decomposition is to apply it directly, similarely to the KSVM algorithm. However doing so suffers from severe numerical unstabilities.
Instead we observe that the proposed decomposition can be applied in a rank-one fashion. This features makes it possible to design a solver that can compute the kernel elements on the fly and produce directly a solution in the RKKS.

\vspace{-0.7em}
\subsection{TrIK-SVM algorithm}
\vspace{-0.7em}
We propose here a dedicated solver that computes at each step both the solution in a RKHS and in the
RKKS, so that gradient steps are performed in the minimization setting and optimality conditions are
checked in the RKKS. The algorithm is based on a active set method \cite{vishwanathan_simplesvm_2003}.
The algorithm iteratively checks dual or primal constraints and updates the solution after each modification of the support vector set.
The dual checking deals with bounds on $\talpha$: in case of a bound violation, the corresponding point
is removed from the set of support vectors. The primal checking deals with the good classification
constraint: in case of a violation, it means that there remains some misclassified training examples
and one is picked to be added to the set of support vectors.
Algorithm \ref{algo:activeset} gives the global scheme in which we insert the indefiniteness treatment. Steps {\bf [1.]} and
{\bf [3.]} are modified while step {\bf [2.]} will not.
In the following, we present in parallel the hard-margin case (HM) and the soft-margin case (SM).
Bounded support vectors are subscripted {\em (bsv)} and free support vectors are subscripted {\em (sv)}.
Matrices $\tG$ and $\G$ are the labeled versions of $\tK$ and $\K$ ($G = Y \K Y$).

\begin{algorithm}
\footnotesize
\label{algo:activeset}
\caption{TrIK-SVM}
\begin{algorithmic}[1]
\Require $y,C,K,\lambda,SV$
\State $converged \leftarrow false$
\While {$not(converged)$}
  \State \bf{[1.]} $[\tilde{\alpha}, \alpha, b] \leftarrow linearSystem (K,C,y,\lambda, SV)$
  \State \bf{[2.]} $[\tilde{\alpha},SV,adm] \leftarrow dualAdmissibility (\tilde{\alpha},SV)$
  \If {$adm$}
    \State \bf{[3.]} $[SV,opt] \leftarrow primalOptimality (\tilde{\alpha},b,K,C,y,\lambda,SV)$
    \If {$opt$}
      \State $converged \leftarrow true$
    \EndIf
  \EndIf
\EndWhile
\State \Return $\alpha,b$
\end{algorithmic}
\end{algorithm}
\vspace{-0.7em}

\noindent {\bf Step [1.]} The idea is to compute at each step both $\talpha$ (already done) and $\alpha$. Originally, this step, restricted to examples that are in the
set of current support vectors, consists in computing $b$ (see \cite{vishwanathan_simplesvm_2003} for details) and $\talpha$.
Introducing our kernel decomposition, and noting that only diagonal terms are different between $\G$ and $\tG$, we obtain that $\alpha_{sv} = (I - 2\lambda \G^{-1}_{sv,sv})\talpha_{sv}$.

\vspace{-0.9em}
\begin{equation}
\label{eq:step1}
\small
\begin{array}{ll}
(HM) & (SM) \\
\tG_{sv,sv}\talpha_{sv}+ b y_{sv} = \mathbf{1}_{sv} &
  \tG_{sv,sv}\talpha_{sv} + C \tG_{sv,bsv} \mathbf{1}_{bsv}+ b y_{sv} = \mathbf{1}_{sv} \\
 \talpha= \tG^{-1}(\mathbf{1}_{sv} - b y_{sv}) &  \talpha= \tG^{-1}(\mathbf{1}_{sv} - C\tG_{sv,bsv} - b y_{sv}) \\
\G_{sv,sv} P_{sv,sv} \talpha_{sv}+ b y_{sv} = \mathbf{1}_{sv} &
 \G_{sv,sv}P_{sv,sv}\talpha_{sv}+ C \G_{sv,bsv} \mathbf{1}_{bsv}+ b y_{sv} = \mathbf{1}_{sv} \\
 \alpha_{sv}= P_{sv,sv}\talpha_{sv} &  \alpha_{sv}= P_{sv,sv}\talpha_{sv}
\end{array}
\end{equation}

\noindent {\bf Step [2.]} This step aims at checking the dual admissibility, ie. checking that all $\talpha_i \geq 0$ and
projecting the solution in the admissible set if it is not. Since in the RKKS $\alpha$ is not constrained \cite{loosli_learning_2016}, this step can only be
performed in the RKHS.

\noindent {\bf Step [3.]} This last step is performed once the current set of support vector gives an admissible
solution in the dual. Its goal is to check the optimality in the primal by verifying that all non support
vectors are well classified, which has to be true in the RKKS.

\noindent Using the modified steps in the original algorithm, we can compute values for $\tK$ and $\K$ on the fly, or
use a cache strategy for more efficiency. Since both solutions are computed on the SV set at each
step, the output $\alpha$ can be sparse.

\vspace{-0.7em}
\subsection{Experimental evaluation of TrIK-SVM}
\vspace{-0.3em}
KSVM has already proven its interest, so the goal of the presented experiments is to check that TrIK-SVM behaves at least as well as KSVM.
We took randomly generated dataset (based on {\texttt make\_blobs} in scikit-learn) in
dimension 5 with 4 centers (2 are assigned to class -1, 2 are assigned to class 1, at random). Resulting
problems may be almost separable and others very overlapping.

\begin{figure}[ht]
\centering
\includegraphics[width=0.52\linewidth]{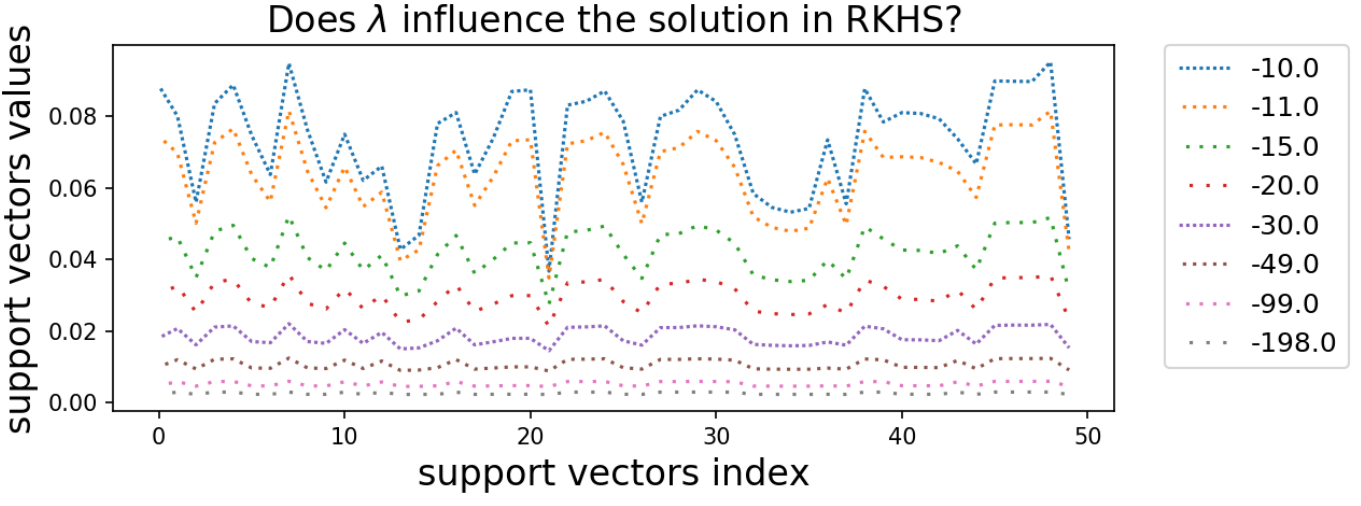}
\includegraphics[width=0.43\linewidth]{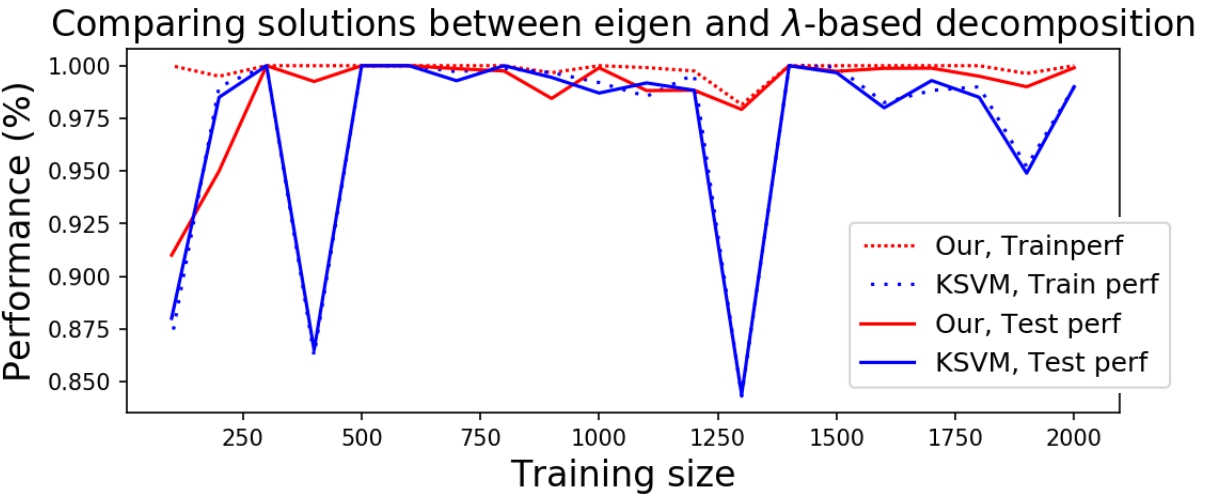}
\includegraphics[width=0.52\linewidth]{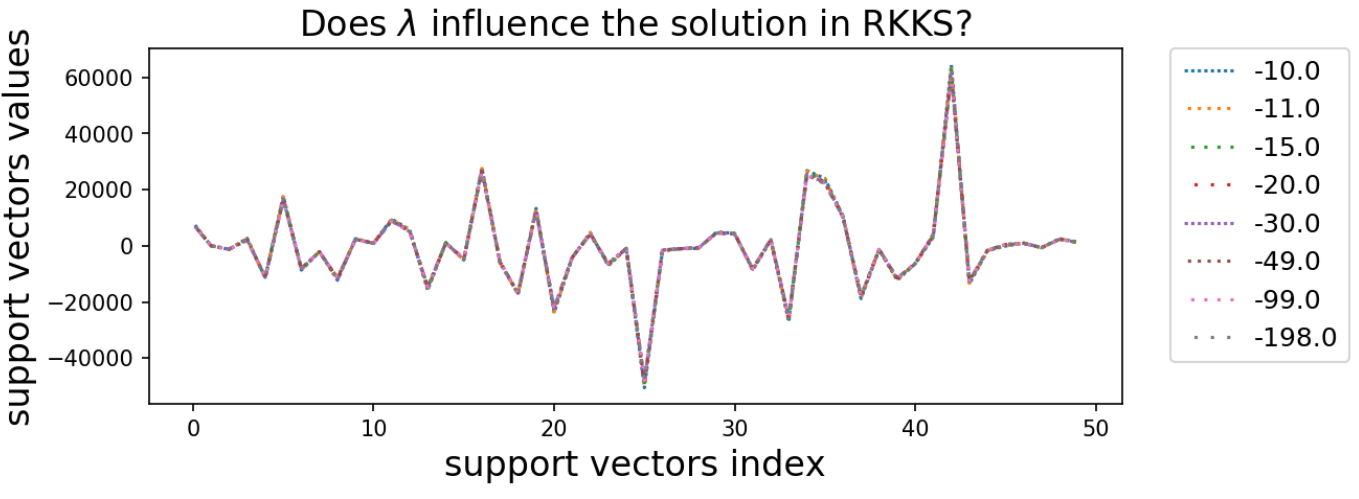}
\includegraphics[width=0.43\linewidth]{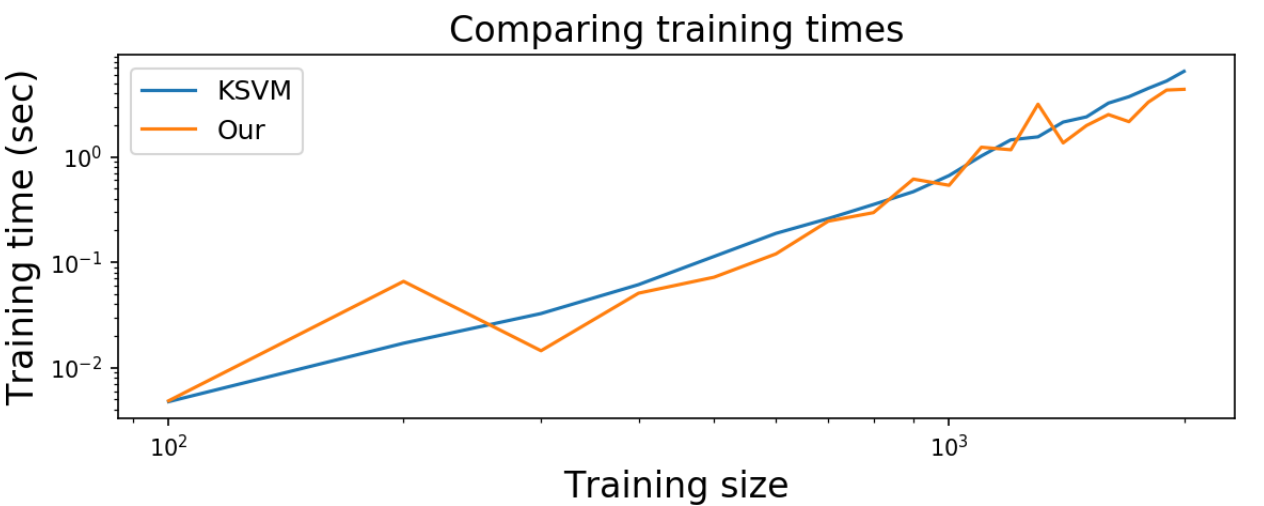}
\vspace{-0.7em}
\caption{\label{fig:1} \footnotesize
{\bf Left:} solutions in the RKHS (top) and in RKKS (bottom) for decreasing $\lambda$ values,
starting from the least computed eigenvalue. We
observe that solutions are similar with a scale factor in RKHS but quasi-identical in the original
RKKS.
{\bf Right top:} train and test accuracy on 30 toy datasets generated randomly, and
of increasing sizes (from 100 to 3000 training examples). On this task, TrIK-SVM
performs at least as well as KSVM.
{\bf Right bottom:} training time in loglog scale.
\vspace{-0.7em}
}
\end{figure}

\vspace{-0.7em}
\paragraph{Sensitivity to $\lambda$}
The first point to check about our decomposition is linked to the fact that we claim
that any value under the least eigenvalue of the spectrum will produce a valid decomposition. If it’s
true on the paper, one might wonder on the numerical effect. Figures \ref{fig:1} illustrate the stability of the
solution in the RKKS independently of $\lambda$. We run TrIK-SVM on one randomly picked dataset, with
different values of $\lambda$, starting from the true least eigenvalue up to 20 times this value. We report the $\alpha$
and $\talpha$ values for each $\lambda$.

\vspace{-0.7em}
\paragraph{TrIKSVM vs KSVM}
The instabilities due to the eigen-decomposition in KSVM are
more easily observed for datasets with many similar training points. In our experiments, it happens
more often than for real datasets, since all training point are picked in a restricted space.
Figure \ref{fig:1} present an illustrative result of the accuracy performance of TrIK-SVM and KSVM. We
observe that TrIK-SVM is competitive over KSVM, which suffers from numerical instabilities. We
also plot on figure \ref{fig:1} the corresponding training time for each, on loglog axis. At first sight the advantage of TrIK-
SVM is not obvious, but we need to mention that here, KSVM is internally using libsvm which is
fully optimized, while the current TrIK-SVM implementation is fully implemented in Python without
any trick and pre-computes the complete kernel, so there’s space for improvement.

%
%
%

\vspace{-0.7em}
\section{Conclusion}
\vspace{-0.7em}
This paper introduces a novel algorithm called TrIK-SVM that solves most of the KSVM limitations:
it provides a sparse solution, that can be computed without pre-computing the full kernel matrix
and thus that can benefit from advanced cache strategies. Since the hyper-parameter $\lambda$ does not
influence the quality of the solution neither the computation time, given that it is sufficiently negative,
TrIK-SVM comes without additional parameter to tune compared to a classical SVM.
Future work includes application to problems others than binary classification, including Multiple Kernel Setting, and a more efficient implementation of the proposed algorithm.


\begin{footnotesize}


\bibliographystyle{unsrt}
\bibliography{bib}

\end{footnotesize}


\end{document}